\documentclass[10pt,twocolumn,letterpaper]{article}
\usepackage{cvpr}
\usepackage{times}
\usepackage{epsfig}
\usepackage{graphicx}
\usepackage{amsmath}
\usepackage{amssymb}
\usepackage{multirow}
\usepackage{float}
\usepackage[breaklinks=true,bookmarks=false]{hyperref}

\cvprfinalcopy 

\def\cvprPaperID{****} 

\begin{document}

\title{U-Net Based Multi-instance Video Object Segmentation}

\author{Heguang Liu\thanks{ Senior software engineer at Uber Technologies, Inc and SCPD student at Stanford University.} \\
Uber Technologies, Inc
\\Stanford University\\
{\tt\small heguangl@uber.com}
\and
Jingle Jiang\thanks{ Ex software engineer at Uber Technologies, Inc and SCPD student at Stanford University at the time of this work. Has since left the company.} \\
Uber Technologies, Inc
\\Stanford University\\
{\tt\small jxj142@gmail.com}
}

\maketitle

\begin{abstract}
Multi-instance video object segmentation is to segment specific instances throughout a video sequence in pixel level, given only an annotated first frame. In this paper, we implement an effective fully convolutional networks with U-Net similar structure built on top of OSVOS fine-tuned layer. We use instance isolation to transform this multi-instance segmentation problem into binary labeling problem, and use weighted cross entropy loss and dice coefficient loss as our loss function. Our best model achieves F mean: 0.467 and J mean: 0.424 on DAVIS dataset, which is a comparable performance with the State-of-the-Art approach. But case analysis shows this model can achieve a smoother contour and better instance coverage, so it's better for recall focus segmentation scenario. We also did many experiments on other convolutional neural networks, including SegNet, Mask R-CNN, and provide insightful comparison and discussion.  
\end{abstract}

\section{Introduction}
Video object segmentation targets at segmenting a specific object throughout a video sequence, given only an annotated first frame ~\cite{Alpher15}. It requires labeling each instances in pixel level accuracy. Multi-instance video object segmentation is  challenging because instances occluding each other often causes failure in tracking. In recent years, it has gained increasing popularity due to its wide usage in autonomous driving vehicles, motion tracking, video summarization, etc. 
\par In this project, DAVIS (Densely Annotated VIdeo Segmentation) 2017 Dataset~\cite{Alpher03} will be used for training and evaluating. Specifically, the input is a sequence of RGB images, and an annotated first frame to indicate the object of interest. The output is the annotation of each instance for each frame. This paper mainly focus on approaches which doesn't use temporal information, to meet the real-time processing need.
\begin{figure}[H]
\includegraphics[width=\linewidth]{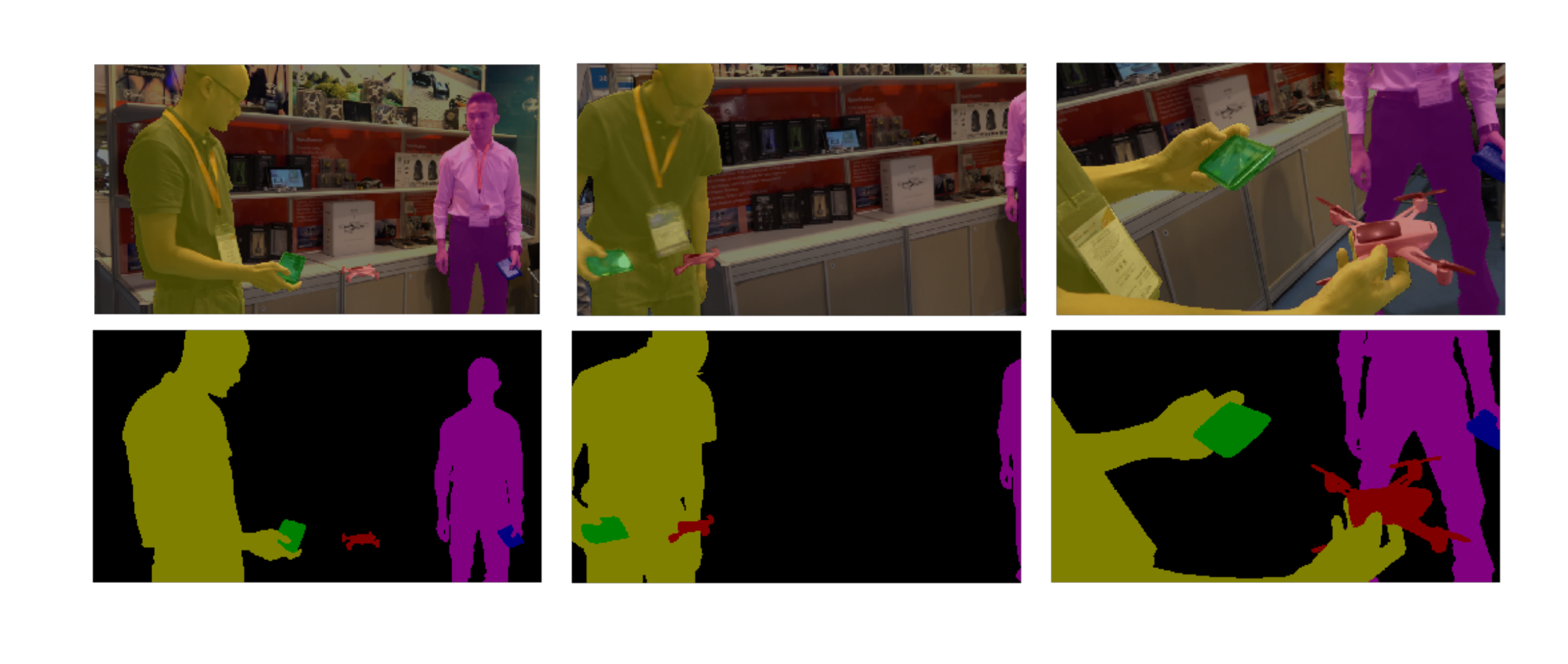}
\caption{Example annotations of DAVIS 2017 dataset}
\end{figure}
\par After experimenting many approaches, including SegNet~\cite{Alpher14}, U-Net and Mask R-CNN~\cite{Alpher13}, we identified a fully convolutional neural networks with an U-Net similar architecture, on top of an OSVOS (One-Shot Video Object Segmentation) ~\cite{Alpher07} fine-tuned layer achieves the best result. This model has a comparable performance with the non-temporal State-of-the-Art OSVOS on DAVIS 2017 dataset. From the case analysis, we found the annotation produced by this model is more complete and with a smoother contour, compared with OSVOS. Thus it's a better model for recall focus scenario whereas OSVOS for contour accuracy focused scenario.
\par This paper also includes insightful discussion about selections on object isolation, loss function, model structure as well as failure analysis. We hope it can be useful for future researcher.
\section{Related Work}
\subsection{Image Segmentation}
\begin{figure*}
\includegraphics[width=\linewidth]{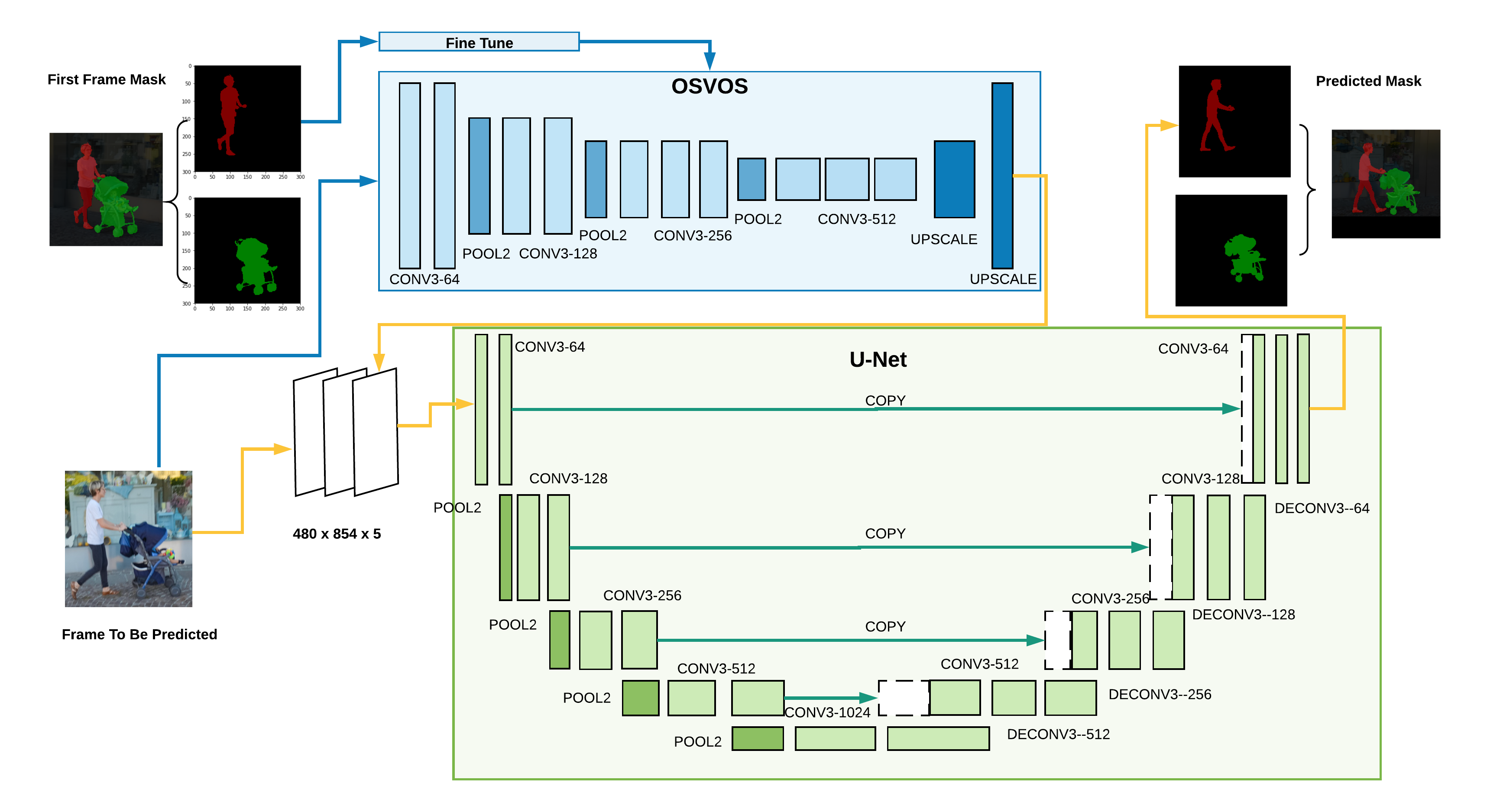}
\caption{U-Net based Fully Convolutional Networks Architecture}
\end{figure*}
\par The State-of-the-Art image semantic segmentation solution typically takes a  encoder-decoder architecture, as popularized in Long, et al's work ~\cite{Alpher12}. Briefly, the encoder is a pre-trained classification network like VGG ~\cite{Alpher18}/ ResNet~\cite{Alpher16}; and decoder's task is to semantically project the discriminative features (lower resolution) learned by the encoder onto the pixel space (higher resolution) to get a dense classification.
\par Few different decoding mechanism has been proposed over the years. When the first breakthrough of semantic segmentation arrives with Long, et al \cite{Alpher12} they only employed a coarse up-sampling for the decoder architecture. Although a good start, the heatmap produced using this technique is quite coarse. Later, SegNet \cite{Alpher14} introduced transposed convolutional layers into the decoder work. Built on similar idea, U-Net \cite{Alpher13} further introduced skip connection to improve.

Another different approach is Region-Based Semantic Segmentation, where a detection is performed first followed by segmentation. It often relies on a two-stage processing: with Region of Interest (ROI) proposal followed by a heavy-lifting working network. Earlier effort on the first stage has been focusing on "segment candidate proposal", e.g. DeepMask~\cite{Alpher11}. But it turns out that these methods are slow and less accurate because segmentation \textit{precedes} recognition. Most recently, He, et al proposed an elegant solution Mask R-CNN ~\cite{Alpher01}. It extends Faster R-CNN by adding a branch for predicting segmentation masks in parallel with the existing branch for classification and bounding box recognition. Mask R-CNN is an end-to-end model, outperforms all existing, single-model entries on the architecture for Pascal VOC~\cite{Alpher05} and MSCOCO ~\cite{Alpher06} challenges.
\subsection{Video Object Segmentation}
There are three leading approaches to Video Object Segmentation, OSVOS~\cite{Alpher07}, MaskTrack~\cite{Alpher08} and Recurrent mask propagation ~\cite{Alpher15}.
\par  The current non-temporal State-of-the-Art is OSVOS. OSVOS approach first converts a pre-trained image classification CNNs, eg VGG-16, to a fully convolutional network by removing the FC layer and insert new loss. It then train this fully convolutional network on the DAVIS dataset. By fine-tuning this network with the first frame of the video, OSVOS generates a one-time model and test it on that entire sequence, using its new weight. This approach shows that we can achieve great result without using temporal information of the video. The model we developed is on top of this approach. 
\par MaskTrack, on the other hand, feed the predicted mask of the previous frame as additional input to the network, to make the input 4 channels. Then train a CNN eg VGG-16, from scratch on a combination of semantic segmentation and image saliency datasets. Finally it adds an identical second stream network, based on optical flow input. Variation approaches like Online Adaptation~\cite{Alpher10} and re-identification~\cite{Alpher09} has also gain great result. 
\par A new released Recurrent Mask Propagation ~\cite{Alpher15} approach formulates a deep recurrent network that is capable of segmenting and tracking objects in video simultaneously by their temporal continuity, so it's able to re-identify them when they re-appear after a prolonged occlusion and achieve great result. This approach uses temporal info, which is not the focus area of this paper. 
\section{Methods}
\subsection{Architecture}
For this paper, we propose an U-Net based fully convolutional networks architecture, as shown in Figure 2. 
\subsection{Key Components}
\subsubsection{Instance Isolation} 
To isolate the instance of the multi-instance mask, we explored 3 approaches:
\begin{itemize}
\item Using the raw multi-instance mask for prediction. The input dimension is H x W x 1, value ranges [0,1,2...N]
\item Project the multi-instance mask to a input of dimension H x W x N, where each layer is the binary labeling of a instance, value ranges [0, 1]
\item Isolate the multi-instance mask image to N binary label inputs, each with dimension H x W x 1, values ranges [0, 1]  
\end{itemize}
After experiments and analysis, we found the third approach is the most effective one. Because it simplifies the multi-instance segmentation problem to a single-instance binary segmentation problem.
\subsubsection{OSVOS}
OSVOS is using a VGG-16 pre-trained on ImageNet ~\cite{Alpher17} as a backbone, removes the FC layer to make it a fully convolutional network, and then trained on DAVIS dataset. After getting this parent model, we use the first frame of the video to tune this model for 500 iterations, which generates a customized model for each video sequence. This fine-tune step turns out to be very important. We used open source OSVOS code to accomplish this step. ~\cite{Alpher07}
\subsubsection{U-Net based Fully Convolutional Networks}
DAVIS dataset only has 4219 training images, so we use a fully convolutional networks, U-Net, to tackle small training data size problem. We first used a series of convolutional layer and max pooling layer to construct a contracting path, in order to capture enough context from the image. Then we used up-sampling layer to replace pooling layer to increase the output resolution. This is a symmetric expanding path of the contracting path. In order to localize, we crop and merge the high resolution feature from the contracting path with these up-sampled output, then send them to successive convolution layer to assemble a more precise output. For this U-Net, we experiment different number of filters in each layer to get the best result.
\subsubsection{Layer: Loss function} 
We mainly explored two types of loss function - weighted cross entropy loss and dice coefficient loss - both of which are typically used for segmentation task. 
\begin{itemize}
\item Weighted Cross Entropy Loss
\textbf{$$\boldsymbol{L=-\sum _{x}\omega(x)\,p(x)\log q(x).}$$}where $p(x)$ denotes the true distribution. $q(x)$ denotes the predicted distribution from the neural network. $\omega(x)$ is a weight coefficient. 
\par Weight coefficient is to scale up the contribution from foreground class when foreground only constitute small area comparing to background. Without the weight, the model would simply predict all pixels as background. 
We set $\omega(x)$ as the ratio of background pixel count and foreground pixel count.
\item Dice Coefficient Loss
\textbf{$$\boldsymbol{L={1-\frac {2|X\cap Y|}{|X|+|Y|}}}$$}where {$|.|$} denotes {$L1$} norm. Dice loss is bound between 0 and 1 where 0 suggests no similarity and 1 suggests completely overlapping. It's a more interpretable loss compared with cross entropy.  
\end{itemize}
During experiment, we observed these two loss functions are positive associated, and weighted cross entropy loss is more numerical stable than dice coefficient loss. So we use the weighted cross entropy loss to train the model and use both during model evaluation.  
\section{Dataset and Metrics}
\begin{figure*}
\includegraphics[width=\linewidth]{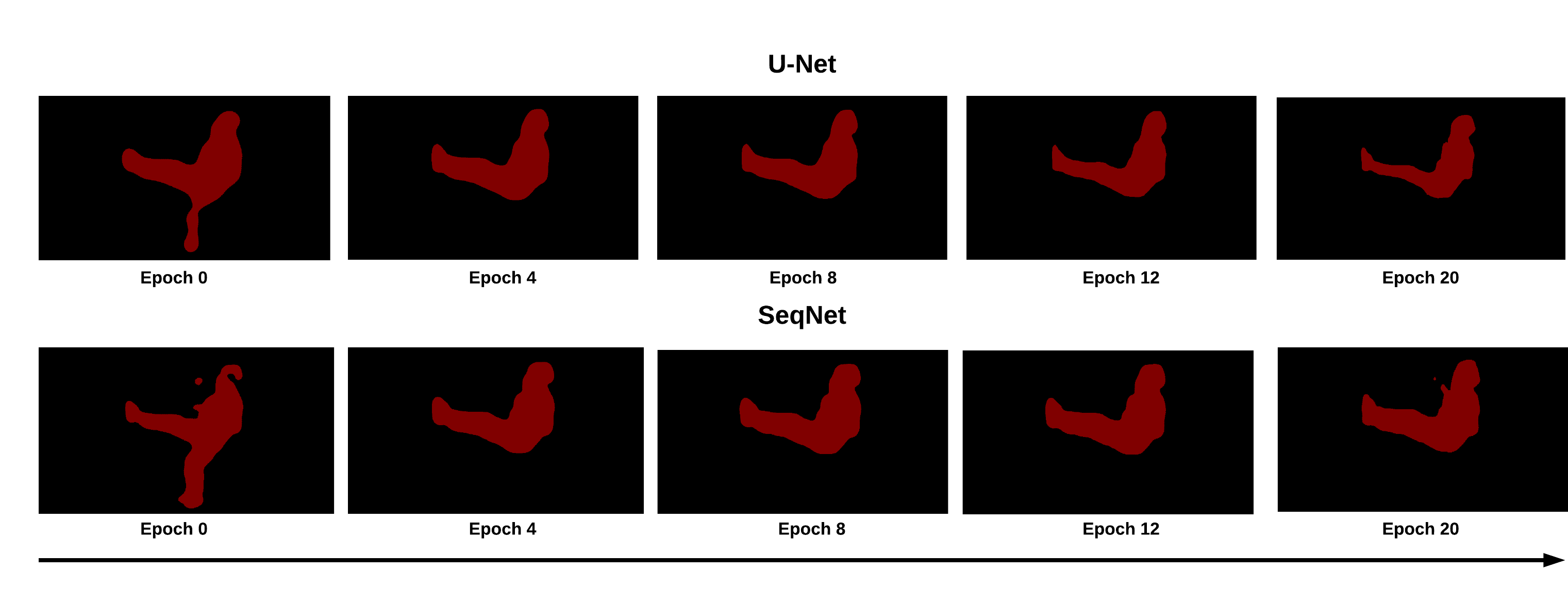}
\caption{Comparison of SegNet and U-Net over training epochs}
\end{figure*}
\subsection{DAVIS Dataset}
DAVIS 2017 Challenge~\cite{Alpher03} on Video Object Segmentation was used for this project. DAVIS dataset spans multiple occurrences of common video object segmentation challenges, such as occlusions, motion blur and appearance changes. On top of DAVIS 2016 dataset ~\cite{Alpher04}, the 2017 dataset added multi-instance challenge. This requires semantic/instance-level segmentation. A detailed summary of the dataset is shown in Table 1. 
\begin{table} [h]
\begin{center}
\caption{DAVIS 2017 Dataset}
\begin{tabular}{llll}
\multicolumn{1}{c}{\bf } &
\multicolumn{1}{c}{\bf Train } &
\multicolumn{1}{c}{\bf Val } &
\multicolumn{1}{c}{\bf Test-Dev}\\ 
\hline \\
Number of sequence & 60 & 30 & 30 \\
\hline
Number of images & 4219 & 2023 & 2037\\
\hline
Mean number of objects & 2.3 & 1.97 & 2.97\\
\hline
\end{tabular}
\end{center}
\end{table}
\subsection{Evaluation Metrics}
We used official evaluation code from DAVIS-2017 to report our model performance. Specifically, we evaluated the following:
\begin{itemize}
\item \textbf{Region Similarity}: the intersection-over union between the predicted mask M and ground-truth G. Note the similarity between this metric and the Dice coefficient loss we introduced earlier. 
\textbf{$$J=\frac{|M\cap{G}|}{|M\cup{G}|}$$}
\item \textbf{Contour Accuracy}: the Harmonic mean of contour's precision and recall
\textbf{$$F=\frac{P_{c}*R_{c}}{P_{c}+R_{c}}$$}
\end{itemize}
\subsection{Training Setup}
The model was built on tensorflow 1.8 framework and implemented using Python 3.6. The code is open-sourced and can be found at \href{$https://github.com/hyuna915/video_segmentation$}{$github.com/hyuna915/video_segmentation$}. In addition to original code to pre-process image, isolate objects, construct U-Net, and SegNet graph, we used the davis-2017 code repository for model evaluation and OSVOS code repository for constructing OSVOS layer.
\par Our model was trained on N1-HighMem-8 instance, with v8CPU, 52 GB memory, on Google Cloud Compute Platform. We've also attached NVIDIA Tesla P100 GPU with 16GB memory to the virtual machine. Due to the sheer amount of model structure/parameters we have experimented upon, in total 5 GPUs are used for this project. 
\section{Experiments and Results}
\subsection{Model Performance Comparison}
We experimented a few different architectures. The results of the two most promising architecture, SegNet and U-Net, are summarized on Table 2. It is observed that best U-Net has a J-mean comparable with the State-of-the-Art OSVOS, in spite of a slightly worse F-mean. Figure 3 shows the convergence pattern of the two architectures on an example training image. 
\begin{table} [H]
\caption{Model Performance Comparison}

\begin{center}
\begin{tabular}{lllll}
\hline
\multicolumn{1}{c}{\bf Model} &
\multicolumn{1}{c}{\bf J Mean } &
\multicolumn{1}{c}{\bf F Mean} &
\multicolumn{1}{c}{\bf Dice Loss } &
\multicolumn{1}{c}{\bf CE Loss } 
 \\ 
\hline \\
OSVOS & 0.499 & 0.592 & - & -   \\
\hline
SegNet & 0.347 & 0.214  & 0.407 & 1.001 \\
\hline
Best U-Net & 0.424 & 0.467 & 0.289 & 1.029 \\
\hline
\end{tabular}
\end{center}
\end{table}

\par Furthermore, a sample annotation comparison between the two is shown on Figure 4. From Figure 4 (c), we noticed U-Net tends to produce a more complete instance with a better coverage and smoother contour in exchange of contour accuracy (i.e. lower F value). As a comparison, OSVOS tends to have less coverage and a rigid contour. We believe our model works better for recall-focus scenario, like pedestrian segmentation. 

\par SegNet, on the other hand, has a much coarse contour with lower precision. Its F-mean is much lower compared with U-Net. The reason is likely to be the absence of merging step, which connects the high resolution features from contracting path with the up-sampling features from the expanding path. Without this connection, SegNet tends to "forget" high-resolution details by the end of the contracting process. We can see this clearly in Figure 4 (d).
\begin{figure}[H]
\includegraphics[width=\linewidth]{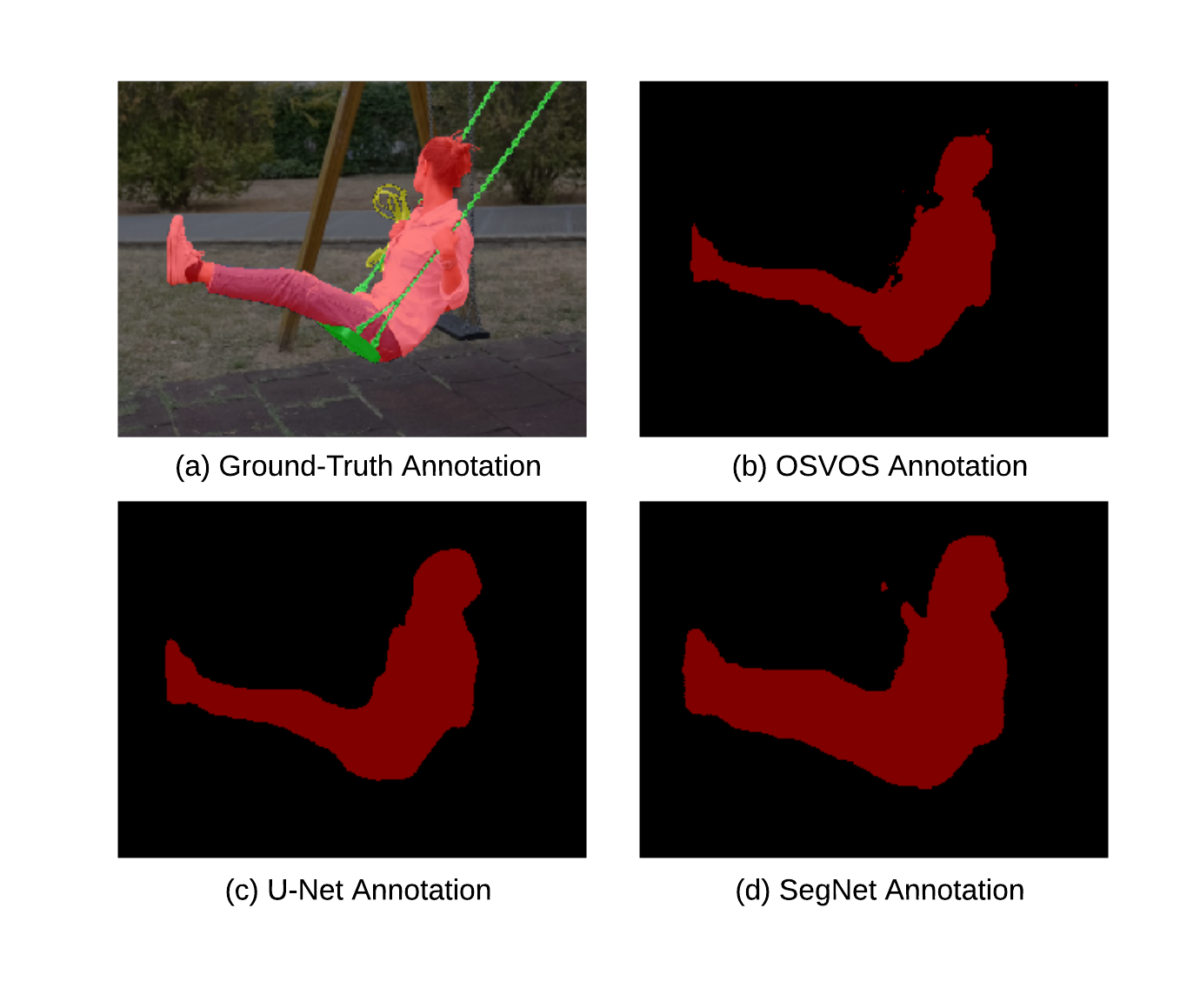}
\caption{Comparison of Annotation Produced by Different Models}
\end{figure}

\subsection{Hyper-parameters Tuning}
For U-Net, we carried many hyper-parameters tuning experiments, including tuning on number of filters in each U-Net layer, batch size and learning rate. The experiment results are summarized in Table 3.
\begin{table} [H]
\caption{U-Net Hyper-parameters Tuning Results}
\begin{center}
\begin{tabular}{llllll}
\hline
\multicolumn{1}{c}{\bf U-Net Filter} &
\multicolumn{1}{c}{\bf J } &
\multicolumn{1}{c}{\bf F } &
\multicolumn{1}{c}{\bf Params} &
\multicolumn{1}{c}{\bf Lr} &
\multicolumn{1}{c}{\bf Batch} \\
\hline
16,32,64 & 0.314 & 0.163 & 700K & 4e-5 & 20 \\
\hline
32,64,128 & 0.345 & 0.325 & 1M & 4e-5 & 20 \\
\hline
64,128,256,512 & 0.419 & 0.430 & 31M & 2e-5 & 8 \\
\hline
64,128,256,512 & 0.424 & 0.467 & 31M & 4e-5 & 8 \\
\hline
\end{tabular}
\end{center}
\end{table}
Figure 5 shows the train loss (first row) and test loss (second row) over 10K iterations for two of our hyper-parameters tuning experiment. the periodic up and downs in training loss is because we were not able to do shuffling on training dataset, due to GPU memory limitation. Although learning rates were different, qualitatively they showed similar trend. The training loss keeps going down while the test loss reaches the best point and goes up afterwards. 
\par Over simplified U-Net with 700K parameters wasn't able to achieve good performance (table 3). The best U-Net Model has convolutional filter number of 64, 128, 256, 512, contains 31M trainable parameters, with learning rate 4e-5 and batch size 8. This model is not over-fitting because both training and test loss is around 1.02. Due to the fact that U-Net has 31M+ parameters in the graph, the batch size can't go beyond 8 to prevent GPU out of memory. The training of the best U-Net takes around 1 hour per epoch and 8 hours to fully converge. 
\begin{figure}[H]
\includegraphics[width=\linewidth]{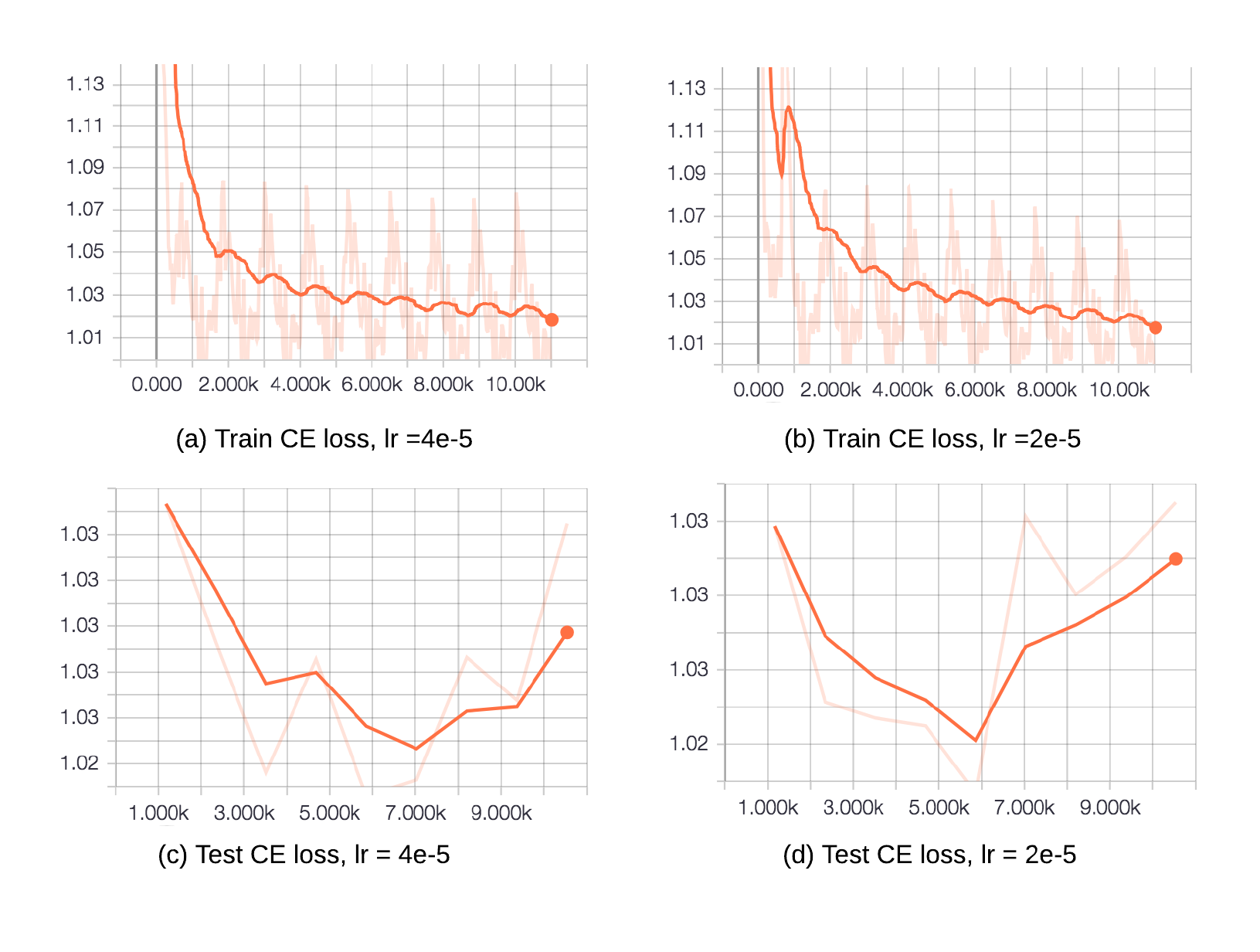}
\caption{Train Loss and Test Loss with different learning rate}
\end{figure}
\subsection{Case Analysis}
We list 4 representative test annotations produced by our best U-Net model, as shown in Figure 6. Based on the case study, we have the following observations:
\begin{enumerate}
\item The model can handle intensive motion and appearance change gracefully.
\par In Figure 6(a) drift chicane sequence, the annotation starts with a small and far car object, which makes multiple turns accompanied by drastic appearance change, distance change, and background fogs. The model keeps good and sharp track on the moving vehicle. 
\item The model can handle multi-instances with similar motion very well, even with overlapping.
\par In Figure 6(b) Horse Jump sequence, the rider and the horse has similar motion. The model can keep track of the two objects and draw relatively clear boundaries.
\item The model doesn't handle multiple object collusion very well.
\par In Figure 6(c) Camel sequence, the annotation is very accurate in the first few frames, but when the target camel bypass another camel, the model starts tracking both camels afterwards. 
\item The model lost track when object goes beyond image boundary and goes back.
\par In Figure 6(d) Motocross jump, the annotation is very accurate in the first few frames, but when the rider moves out of the image boundary, the model lost track of the rider and start tracking only the motorcycle afterwards.   
\end{enumerate}
\begin{figure*}
\begin{center}
\includegraphics[width=1.1\linewidth]{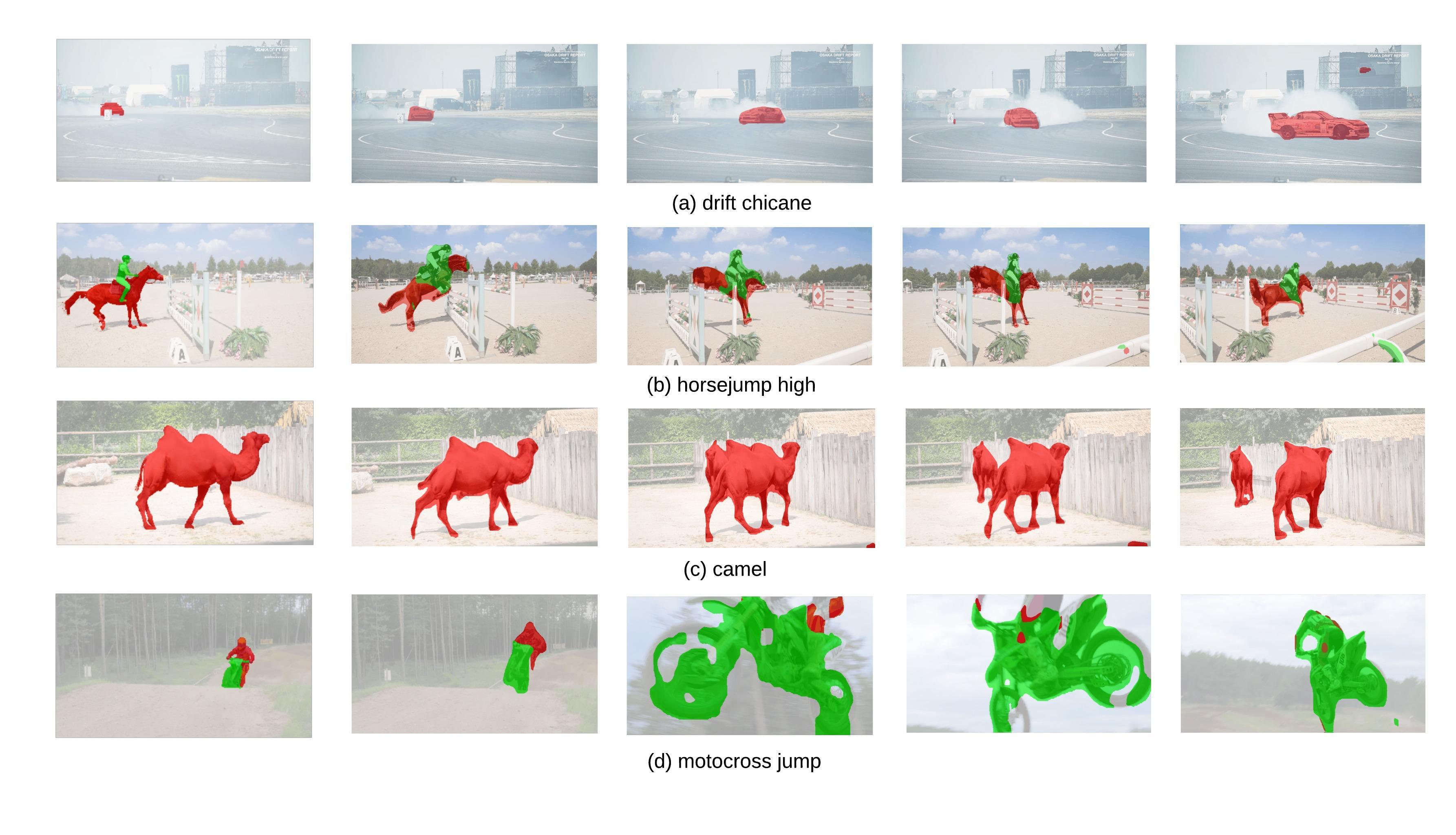}
\caption{U-Net Predicted Annotations on DAVIS 2017 Test Set}
\end{center}
\end{figure*}

\subsection{Failed Experiments and Discussions}
For this paper, we also explored few other approaches, which did not achieve a good result. We still want to report them and hope it can help with future research.     
\begin{itemize}
\item \textbf{Mask R-CNN}: 
We noticed Mask R-CNN didn't produce reasonable result for semi-supervised video segmentation problem as in the case of DAVIS-2017 dataset, for two reasons. First, being a instance segmentation algorithm, Mask R-CNN requires \textit{recognition} for \textit{segmentation}. This caused difficulty  when DAVIS dataset tracks object that Mask R-CNN model has not seen before. Figure 7 is a classic example where Mask R-CNN, trained with COCO dataset that did not include camel, was used to segment camel sequence. Consequently, Mask R-CNN mistakingly predict the camel as an horse with 0.998 confidence and further tried to correct the shape to be a horse, hurting the annotation accuracy.
\begin{figure}[H]
\begin{center}
\includegraphics[width=0.5\linewidth]{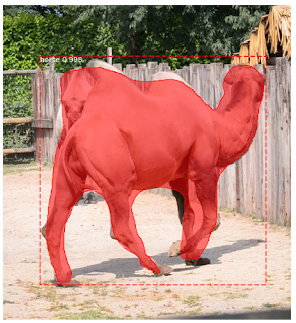}
\caption{Annotation Produces by Mask R-CNN}
\end{center}
\end{figure}
\par Another failure reason on vanilla Mask R-CNN is it was not able to incorporate the semi-supervised information provided on the first image. This often lead to unnecessary objects being segmented in the following frames, as well as wrong segmentation orders for the original targets. However, this problem may be mitigated by training the Mask R-CNN backbone on the first image of each video sequence, similar as OSVOS's approach. 

\item \textbf{Unweighted Cross Entropy as Loss Function}: 
For DAVIS dataset, since most of the instance is very small compared with the background, when we tried to use unweighted cross entropy as loss function, almost all the pixels are predicted as background.   
\item \textbf{Direct Feed Multi-instance Image}:
We also tried to feed multi-instance image to the networks directly and transform this problem into a multi-class classification problem. Because convolutional networks is not able to project the value of 1, 2, 3 to layer 1, 2, 3. The result produced by this approach is also very poor. 
\end{itemize}
\section{Conclusion}
For this paper, we implement and compare a number of fully convolutional networks to tackle the multi-instances video object segmentation problem. Among SegNet, U-Net, Mask R-CNN, U-Net based architecture achieves the best result with \textbf{F mean: 0.467} and \textbf{J mean: 0.424} . This result is comparable to the current State-of-the-Art approach on DAVIS Dataset. Based on the model comparison, we found the annotation produced by this model is more complete and with a smoother contour, compared with OSVOS. Thus it works better under recall focus scenario. 
\par From the case study, we noticed this model doesn't perform well on 2 cases: 1). Multi-instance occlusion 2). Instance lost tracking after it goes out of image boundary and goes back. In the future, we could try 1) Recurrent Neural network to better tracking each object by its temporary continuity to handle occlusion. 
2) Adaptive object re-identification to prevent target lost. 
\section{Acknowledgements}
We would like to thank Professor FeiFei Li, Instructor Justin Johnson and Serena Yeung
and all the TAs for the great class experience and project setting. Also want to thank Google Cloud for sponsoring GPU instances for model training. 
{\small
\bibliographystyle{ieee}
\bibliography{egbib}

\begin{thebibliography}{10}\itemsep=-1pt

\bibitem{Alpher14}
V.~Badrinarayanan, A.~Kendall, and R.~Cipolla.
\newblock Segnet: A deep convolutional encoder-decoder architecture for image
  segmentation.
\newblock {\em CVPR}, 2015.

\bibitem{Alpher07}
S.~Caelles, K.-K. Maninis, J.~Pont-Tuset, L.~Leal-Taixe, D.~Cremers, and L.~V.
  Gool.
\newblock One-shot video object segmentation”.
\newblock {\em CVPR}, 2017.

\bibitem{Alpher05}
M.~Everingham, L.~V. Gool, C.~K.~I. Williams, J.~Winn, and A.~Zisserman.
\newblock The pascal visual object classes (voc) challenge.
\newblock {\em Int J Comput Vis DOI 10.1007/s}, 2009.

\bibitem{Alpher01}
K.~He, G.~Gkioxari, P.~Dollár, and R.~Girshick.
\newblock Mask r-cnn.
\newblock {\em CVPR}, 2017.

\bibitem{Alpher18}
A.~Krizhevsky, I.~Sutskever, and G.~E. Hinton.
\newblock Very deep convolutional networks for large-scale image recognition.
\newblock {\em CVPR}, 2015.

\bibitem{Alpher17}
A.~Krizhevsky, I.~Sutskever, and G.~E. Hinton.
\newblock Imagenet classification with deep convolutional neural networks.
\newblock {\em NIPS}, 2017.

\bibitem{Alpher16}
X.~Li, Y.~Qi, and Z.~Wang.
\newblock Video object segmentation with re-identification.
\newblock {\em CVPR}, 2017.

\bibitem{Alpher09}
X.~Li, Y.~Qi, Z.~Wang, K.~Chen, Z.~Liu, J.~Shi, P.~Luo, C.~C. Loy, and X.~Tang.
\newblock Video object segmentation with re-identification.
\newblock {\em CVPR}, 2017.

\bibitem{Alpher06}
T.-Y. Lin, M.~Maire, S.~Belongie, L.~Bourdev, R.~Girshick, J.~Hays, P.~Perona,
  D.~Ramanan, C.~L. Zitnick, and P.~Dollar.
\newblock Microsoft coco: Common objects in context.
\newblock {\em ECCV}, 2014.

\bibitem{Alpher12}
J.~Long, E.~Shelhamer, and T.~Darrell.
\newblock Fully convolutional networks for semantic segmentation.
\newblock {\em CVPR}, 2016.

\bibitem{Alpher08}
F.~Perazzi, A.~Khoreva, R.~Benenson, B.~Schiele, and A.SorkineHornung.
\newblock Learning video object segmentation from static images.
\newblock {\em CVPR}, 2017.

\bibitem{Alpher04}
F.~Perazzi, J.~Pont-Tuset, B.~McWilliams, L.~V. Gool, M.~Gross, and
  A.~Sorkine-Hornung.
\newblock A benchmark dataset and evaluation methodology for video object
  segmentation.
\newblock {\em CVPR}, 2016.

\bibitem{Alpher11}
P.~O. Pinheiro, R.~Collobert, and P.~Dollar.
\newblock Learning to segment object candidates.
\newblock {\em CVPR}, 2016.

\bibitem{Alpher03}
J.~Pont-Tuset and F.~Perazzi.
\newblock The 2017 davis challenge on video object segmentation.
\newblock {\em CVPR}, 2017.

\bibitem{Alpher13}
O.~Ronneberger, P.~Fischer, and T.~Brox.
\newblock U-net: Convolutional networks for biomedical image segmentation.
\newblock {\em CVPR}, 2015.

\bibitem{Alpher10}
P.~Voigtlaender and B.~Leibe.
\newblock Online adaptation of convolutional neural networks for video object
  segmentation.
\newblock {\em BMVC}, 2017.

\bibitem{Alpher15}
C.~C.~L. Xiaoxiao~Li.
\newblock Video object segmentation with joint re-identification and
  attention-aware mask propagation.
\newblock {\em CVPR}, 2018.

\end{thebibliography}
}
\end{document}